\documentclass[sigconf]{acmart}

\usepackage{multirow}

\usepackage{algorithm}
\usepackage{algorithmic}
\usepackage{subfigure}
\usepackage{amsmath}
\usepackage{amsmath}

\usepackage{newfloat}
\usepackage{listings}

\AtBeginDocument{%
  \providecommand\BibTeX{{%
    \normalfont B\kern-0.5em{\scshape i\kern-0.25em b}\kern-0.8em\TeX}}}

\setcopyright{acmlicensed}
\copyrightyear{2024}
\acmYear{2024}
\acmDOI{XXXXXXX.XXXXXXX}

\acmConference[Conference acronym 'XX]{Make sure to enter the correct
  conference title from your rights confirmation emai}{June 03--05,
  2024}{Woodstock, NY}
\acmISBN{978-1-4503-XXXX-X/18/06}
\usepackage{lineno}

\begin{document}
\nolinenumbers
\title{ Neural Graph Matching for Video Retrieval in Large-Scale Video-driven E-commerce}

\author{Houye Ji}
\affiliation{%
  \institution{JD.COM}
  \streetaddress{P.O. Box 1212}
  \country{China}
  \postcode{43017-6221}
}

\author{Ye Tang}
\affiliation{%
	\institution{JD.COM}
	\streetaddress{P.O. Box 1212}
	\country{China}
	\postcode{43017-6221}
}

\author{Zhaoxin Chen}
\affiliation{%
	\institution{JD.COM}
	\streetaddress{P.O. Box 1212}
	\country{China}
	\postcode{43017-6221}
}

\author{Lixi Deng}
\affiliation{%
	\institution{JD.COM}
	\streetaddress{P.O. Box 1212}
	\country{China}
	\postcode{43017-6221}
}

\author{Jun Hu}
\affiliation{%
	\institution{National University of Singapore}
	\streetaddress{P.O. Box 1212}
	\country{China}
	\postcode{43017-6221}
}

\author{Lei Su}
\affiliation{%
	\institution{JD.COM}
	\streetaddress{P.O. Box 1212}
	\country{China}
	\postcode{43017-6221}
}



	\begin{abstract}

	{
		
		With the rapid development of the short video industry, traditional e-commerce has encountered a new paradigm, video-driven e-commerce, which leverages attractive videos for product showcases  and provides both video and item services for users.
		Benefitting from the dynamic and visualized introduction of items, video-driven e-commerce has shown huge potential in stimulating consumer confidence and  promoting sales.
		In this paper, we focus on  the video retrieval  task, facing the following challenges: (1)  How to handle the heterogeneities  among users, items, and videos?  (2) How to mine the complementarity  between items and videos for  better user understanding?
		In this paper, we first leverage the dual graph to model the co-existing of user-video and user-item interactions in video-driven e-commerce and innovatively reduce user preference understanding to a graph matching problem. 
		To solve it, we further propose  a novel bi-level Graph Matching Network (GMN),  which mainly consists of node- and preference-level   graph matching.
		Given a user, node-level  graph matching aims to match videos and items, while preference-level graph matching aims to match multiple user preferences extracted from both videos and items.
		Then the proposed GMN can generate and improve user embedding by aggregating matched nodes or preferences from the  dual graph in a bi-level manner.
		Comprehensive  experiments show the superiority of the proposed GMN with significant improvements over   state-of-the-art approaches (e.g., AUC+1.9\% and  CTR+7.15\%). 
		We have developed it on a well-known video-driven  e-commerce platform, serving hundreds of millions of users every day.
		
	}
	
\end{abstract}




\received{20 February 2007}
\received[revised]{12 March 2009}
\received[accepted]{5 June 2009}

\maketitle

\section{Introduction}



\begin{figure}
	\centering
	\includegraphics[width=0.995\columnwidth, height=5.5cm]{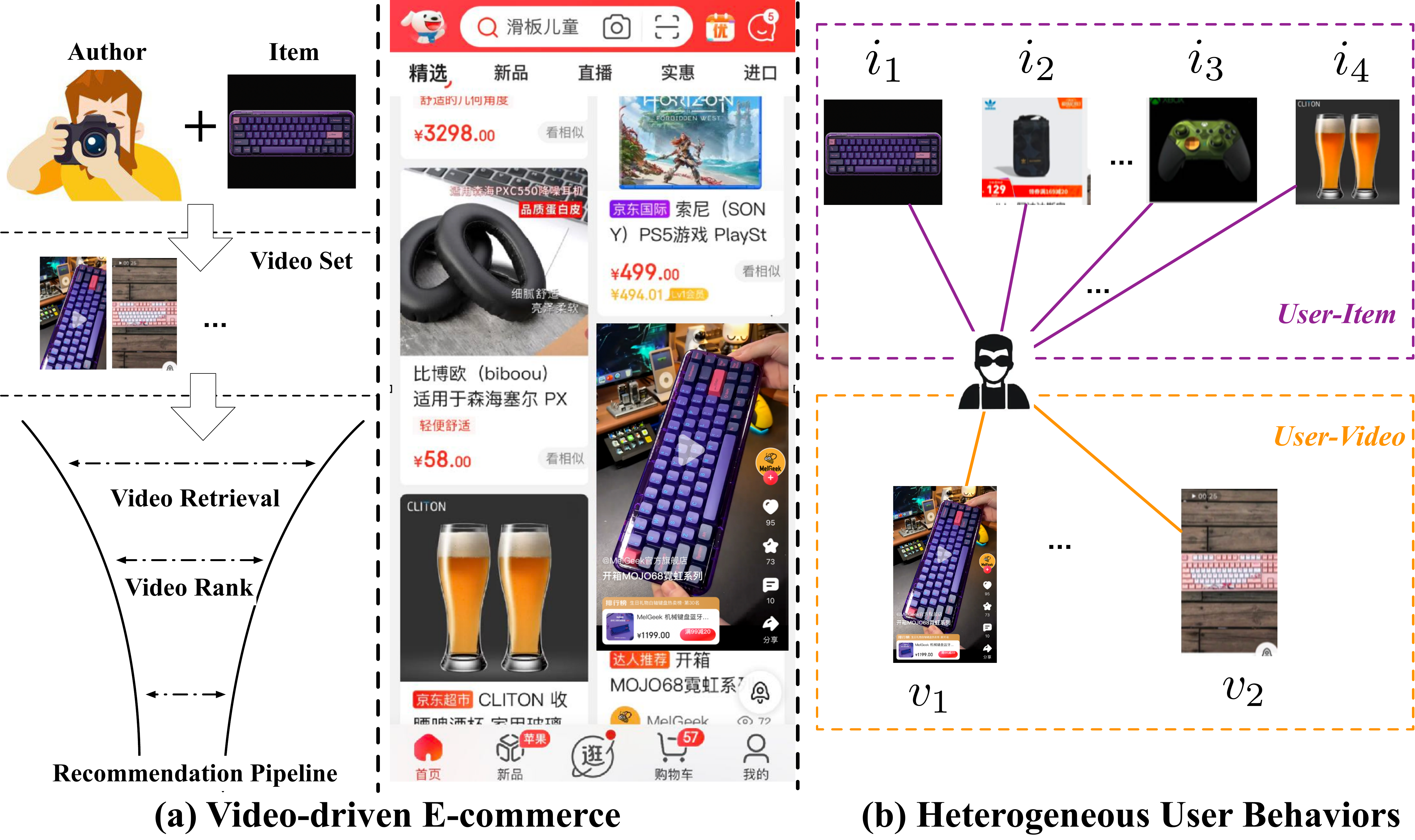}
	\caption{An     example of video-driven e-commerce.}
	\label{fig_demo}
\end{figure}

	In the era of the information explosion, the recommender system has become the most effective way to help users  discover what they are interested from enormous data.
	As two basic Internet applications, e-commerce and  content recommendation  both  provide  the recommender service but focus on either item recommendation or content (e.g., video and live) recommendation.  
	Recently, with the thriving of online applications, there is a surge of \textbf{video-driven  e-commerce} \cite{22cikm_gift},  which integrates  video    and e-commerce for better  services. Benefiting from   vivid and  attractive video shows, video-driven  e-commerce provides a new business paradigm which improves user stickiness and activeness, and has become a new driving force for e-commerce development.
	For example, TikTok and Kuaishou integrate e-commerce with videos, while
	Taobao and  JD.COM     leverage videos   to   improve the user experience in traditional e-commerce.   As shown in Figure \ref{fig_demo}(a),
	JD.COM first incentivizes \emph{authors} to produce \emph{videos}  for related \emph{items},  and then conducts video recommendations for users.
	Due to the large-scale video set,
	industrial recommendation systems usually adopt the classical two-stage architecture: retrieval and ranking \cite{22kdd_ccdr}.
	Given a user, the \emph{retrieval} stage  aims to find thousands of candidate videos from the whole video set, then  the  \emph{ranking} stage aims to  precisely select   hundreds of videos from the retrieved candidate videos.
	In this paper, we focus on the video retrieval stage with the following challenges:
\emph{(1) How to  handle the heterogeneities   among users, items, and videos?}
The characteristics of items (e.g., price, sales, and shop) are markedly different from videos (e.g.,  resolution, duration, and author) and them aim to stimulate user consumption and attract users stay, respectively.
This disparity compels us to treat user-item interactions and user-video interactions with distinct considerations, given the unique characteristics each type of interaction manifests.
\emph{(2) How to mine the complementarity  between items and videos for better user understanding?}
While at a cursory glance these interactions may seem discrete, they exhibit a strong complementary nature upon closer examination. As exemplified in Figure \ref{fig_demo}, when a user interacts with both an item $i_1$ symbolizing a keyboard and a corresponding video $v_1$, it reinforces the user's profound interest in keyboards. In a contrasting scenario, if a user interacts with a beer item $i_4$ but directs their video interactions exclusively towards devices, such as keyboards, it suggests a potential superficial interest in the beer item.

%

%
Based on the above analysis, when  performing video retrieval  in video-driven e-commerce, we need to address the following new requirements:
\begin{itemize}
	\item \textbf{Heterogeneous Interaction Modeling.}  
	Different from traditional e-commerce, the video-driven e-commerce consists of complex heterogeneous interactions among users, videos, and items.
	Intuitively, user interactions on videos and items will show similar patterns and preferences. 
	As shown in Figure \ref{fig_demo}(b), user $u_1$ who wants to buy a keyboard will watch related videos $v1,v2$ and   click item $i_1$.
	How to model the co-existing of heterogeneous interactions is the basis of user preference understanding and video retrieval.
	\item  \textbf{Node-level Graph Matching.} 
	%
	As shown in Figure \ref{fig_demo}(b), video $v_1$ and item $i_1$ are both keyboard-related (high relevance) while video $v_1$ and item $i_4$ are belonging to different categories (low relevance).
	So, when jointly modeling heterogeneous interactions for user preference understanding, it is necessary to properly distinguish and match the relevance between videos and items.
	\item \textbf{Preference-level Graph Matching.} 
	Intuitively, diverse user preferences extracted from both videos and items are similar to each other.
	As shown in Figure \ref{fig_demo}(b), items $\{i_1, i_2, i_3 \}$  indicate user preference on electronic products, which is similar to user preference on video $\{v_1, v_2\}$.
	But, item $\{i_4\}$  indicates user preference on food, which is different from his video preference.
	It is necessity to mine the  the differences and relevances for user preference understanding.
	Particularly, how to match multiple user preferences on both videos and items is still an open problem.
\end{itemize}

In this paper, we study a newly emerging  paradigm video-driven e-commerce,  and then  formulate it as the dual graph to model the co-existing of users, videos, and items as well as interactions among them. 
After that, we reduce the user preference understanding  in the dual graph to a graph matching problem. To solve it, we propose a novel \textbf{G}raph \textbf{M}atching \textbf{N}etwork (GMN), which mainly consists of node- and preference-level graph matchings, to improve user preference understanding. 
The contributions of this work are summarized as follows:

\textbullet ~ To our knowledge,  this  is the first time that  leveraging the dual graph  to model  the newly emerging video-driven e-commerce,   and innovatively reduce  the user preference understanding  problem to a graph matching problem.

\textbullet ~ To solve this, we propose a novel  Graph Matching  Network,  called GMN, which mainly consists of node- and preference-level   graph matching. Specifically,   node-level   graph matching aims to mine the potential   relevance between  videos and items  in the dual graph.  Furtherly,    preference-level graph matching further captures the  relevance of user preferences extracted from both  videos and items.


\textbullet ~ Extensive experimental results on both offline and online data demonstrate the superiority of the proposed GMN. Currently, it has been deployed on one of the largest video-driven e-commerce, affecting hundreds of millions of users every day.

\section{Related Work}
\textbf{Recommendation System} (RS), which mainly focuses  on  user preference understanding and recommends diverse types of candidates (e.g., items and videos) to them, 
has accompanied  diverse paradigms, such as  item recommendation \cite{,20sigir_lightgcn,mcrec,23www_autocf}, video  recommendation \cite{22cikm_gift,23sigir_lightgt,22sigir_cmi}, and business-driven recommendation  \cite{21aaai_hgsrec,21www_comb_k}. 
Real-world    RS usually consists of two stages: retrieval   \cite{16recsys_youtube_dnn,19cikm_mind,22kdd_ccdr}
and   ranking   \cite{21sigir_surge,22cikm_gift}.
YoutubeDNN \cite{16recsys_youtube_dnn} retrieves videos for online video services, while MIND~\cite{19cikm_mind} retrieves items for e-commerce.
Unlike them, 
we study a newly emerging video-driven e-commerce, which models the co-existing of items and videos.

\textbf{Cross-domain Recommendation} (CDR) attempts to learn useful knowledge from the source domain   to help the target domain. 
Hu et al.  \cite{18cikm_conet} propose CoNet  to perform 
dual transfer at the unit level.   \cite{19cikm_graphnet,dagcn,22cikm_gift,22kdd_ccdr} unify both source and target domains as
graph-structured data  and perform CDR via graph learning.
Significantly different from the above works, we 
we first model video-driven e-commerce 
as a dual graph and reduce it to a graph matching problem.

\textbf{
Graph Neural Networks}  (GNNs) generalize deep learning to graph-structured data, where graph convolution \cite{17iclr_gcn} aims to aggregate  neighbors information  and update node embedding,
while graph pooling
\cite{18nips_diffpool,20iclr_gmn}  aims to aggregate all node to update graph embedding.
Recently, many works \cite{21sigir_surge,22cikm_gift,22kdd_ccdr} apply GNNs to      recommendation scenarios. 
SURGE~\cite{21sigir_surge} reduces   user behaviors into graphs   and then extracts users' core interests via cluster-aware   graph pooling. 
Furthermore, several works \cite{22cikm_gift,22kdd_ccdr,22www_mgfn} generalize GNN to multiple interactions (or domains).
MGFN~\cite{22www_mgfn} proposes a multi-graph learning for cross domain video recommendation.
%
Different from the above works, we first propose a novel bi-level graph matching network  to align  user behaviors and  preferences on both videos  and items    for   retrieval.

\section{Preliminaries}

\begin{definition}
	\emph{\textbf{Dual   Graph}}. 
The dual graph,  termed as $\mathcal{G}=(\mathcal{G}^{U-V},\mathcal{G}^{U-I})$, describes the co-existing of  user  behaviors on both videos and items in video-driven e-commerce. Here  $\mathcal{N}=\mathcal{N}^{U} \cup \mathcal{N}^{V}\cup \mathcal{N}^{I}$ is the union set  of  $User$, $Video$, and $Item$,  respectively. Also,   $\mathcal{E}=\mathcal{E}^{U-V} \cup \mathcal{E}^{U-I}$ is the edge set, where	$\mathcal{E}^{U-V}=\langle\mathcal{N}^{U}, \mathcal{N}^{V} \rangle$ denotes $User$-$Video$ interaction  and $\mathcal{E}^{U-I}=\langle\mathcal{N}^{U}, \mathcal{N}^{I} \rangle$ denotes $User$-$Item$ interaction. 
\end{definition}

\textbf{Example.}~
Figure \ref{fig_demo}(b) gives an illustrative example of the dual graph in video-driven e-commerce.
Given  a user $u$ and the  dual graph  $\mathcal{G}$, we can obtain his neighbor sets, including $ \mathcal{N}_u^{U-V}=\{ v|\langle u, v \rangle \in \mathcal{E}^{U-V}\} $ from $\mathcal{G}^{U-V}$ and
$ \mathcal{N}_u^{U-I}=\{ i|\langle u, i \rangle \in \mathcal{E}^{U-I}\} $ from $\mathcal{G}^{U-I}$.
In the dual graph, user $u_1$'s  neighbor sets  are  $\mathcal{N}_{u_1}^{U-I}= \{i_1,  i_2, i_3, i_4 \}$
and   $\mathcal{N}_{u_1}^{U-V}= \{v_1, v_2  \}$, respectively.
Furthermore, we also  extract user-centric  sub-graphs  for user $u$  from the dual graph, termed as    $ \mathcal{G}_u=(\mathcal{G}_u^{U-V},\mathcal{G}_u^{U-I})$.  
In detail, $\mathcal{G}_u^{U-V}$  contains multi-hops neighbors of user $u$ and himself (i.e., $ \mathcal{V}_u^{U-V} = \mathcal{N}_u^{U-V} \cup u $), as well as induced edges $ \mathcal{E}_u^{U-V}$. 
Note that,  dual graphs can be widely used in many other related fields, such as  the co-existing of user-music and user-movie, or even  the co-existing of user-item(old/warm domain) and user-item(new/cold domain).


Finally, we formulate the video retrieval problem in video-driven e-commerce as follows:

\begin{equation}
\arg\max_{v \in \mathcal{N}^{V}} Pr(v| u; \mathcal{G}), 
\end{equation}
which aims to retrieve candidate videos $v$ from the whole video set $\mathcal{N}^{V}$ for user $u$ based on the dual graph $ \mathcal{G}$.

%
%
%
%

\begin{figure*}

\centering
\includegraphics[width=1.999\columnwidth, height=6.4cm]{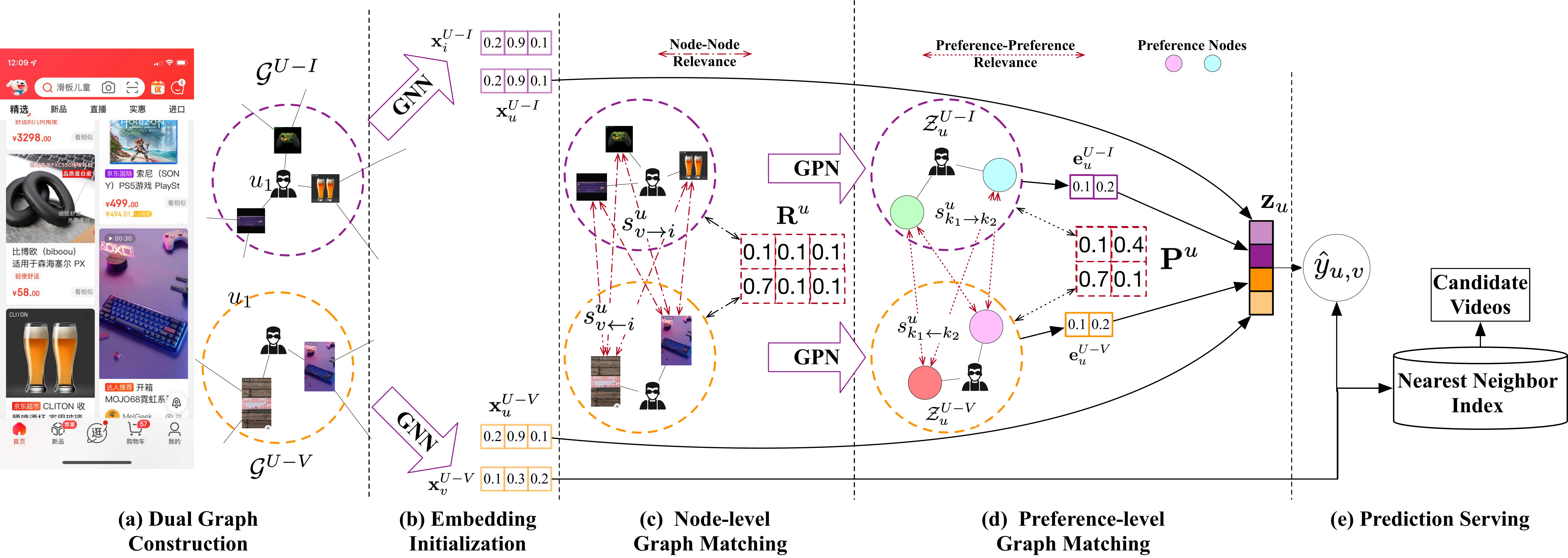}
\caption{The overall framework
of the proposed GMN. (a) Dual graph construction for video-driven e-commerce. (b)  Initialization of embedding through feature embedding and graph neural network. (c) Node-level graph matching. 
(d) Preference-level graph matching. (e) 
Prediction and serving.}
\label{fig_model}
\end{figure*}

\section{The Proposed Model}

In this section, we present a novel  \textbf{G}raph \textbf{M}atching  \textbf{N}etwork, which aims the  to match the users’ behaviors as well as preferences via both node- and preference-level graph matching.
The overall framework of GMN is 
shown in Figure~\ref{fig_model}.

\subsection{Embedding Initialization}

In this section, we first initialize the node embeddings of users, videos, and items via   feature embeddings, and then update them via GNNs.
%
Taking   embedding initialization   of video $v$ as an example,  we  have:
%
\begin{equation}
\mathbf{x}_{v}= \big|\big|_{f} \mathbf{x}_{v}^{f},
\end{equation}
where $||$ denotes the vector concatenation, $	\mathbf{x}_v$ and $\mathbf{x}_{v}^{f}$ denote the initial embedding of video $v$ and the  $f$-th feature embedding of video $v$, respectively. 
After that, we further utilize two interaction-specific    GNNs (i.e., $GNN^{U-V}$ and $GNN^{U-I}$) to update corresponding embeddings.
%
Given  video $v$ and $ \mathcal{G}^{U-V}$,  $GNN^{U-V}$ is able to update the video $\mathbf{x}_v^{U-V}$ embedding via aggregating its neighbors $\mathcal{N}_v^{U-V}$, as follows:
\begin{equation}
\mathbf{x}_v^{U-V} = GNN^{U-V}(\mathbf{x}_u| u \in \mathcal{N}_v^{U-V}).
\end{equation}

Note that,   we focus on bi-level graph matching and 
simply instantiate  $GNN^{U-V}$  as the  $MeanPooling$ \cite{21aaai_hgsrec,21www_comb_k}. 
Similarly,  the embedding of item $i$ via  $GNN^{U-I}$ is $\mathbf{x}_i^{U-I}$.
For user $u$, we can also get his embeddings on the  dual graph    $\mathbf{x}_u^{U-V}$ and  $\mathbf{x}_u^{U-I}$.
%

{ 
%

}

Until now, 
a method way for
video retrieval   is  
to fuse user preferences on   videos and items $i.e., \mathbf{x}_{u}^{U-V}|| \mathbf{x}_{u}^{U-I} $, and then retrieve top videos based on the similarity of their embeddings, 
suffering from  the following weaknesses:
(1) In video-driven e-commerce,  video production  actually revolves around items   and  aims to promote the item purchase. Although videos and items are different types of nodes, there exists potential relevance between them.
(2) It only models each user as a single node embedding while fails to explicitly capture the  graph structure  surrounding users, which  also provides  valuable information for understanding user preferences. So, it is necessary to  encode the user-centric sub-graph as a whole for video retrieval.

\subsection{Node-level Graph Matching}

In this section, we propose node-level graph matching to mine the  relevance between videos and items in the dual graph,  and then enhance their connections via  aggregating  cross-graph neighbors based on    the  matching scores.

To handle large-scale video-driven e-commerce, we 
first extract user-centric  sub-graphs (i.e.,  $\mathcal{G}_u^{U-V}$ and $\mathcal{G}_u^{U-I}$) for user $u$, 
and then actually perform the node-level graph matching $GM_{NL}$ on them, shown as follows:

\begin{equation}
\mathbf{R}^u_{} = GM_{NL}(\mathcal{G}_u^{U-V},\mathcal{G}_u^{U-I}).
\end{equation}
Here   $\mathbf{R}^u  \in \mathbb{R}^{|\mathcal{V}_u^{U-V}| \times |\mathcal{V}_u^{U-I}|} $ 
denotes the  personalized relevance   matrix of user $u$ in   node-level.
Given a   user $u$, 
the  personalized   relevance   score $ r^u_{v,i}$  between   video $v \in \mathcal{G}_u^{U-V}$ and   item $i \in \mathcal{G}_u^{U-I}$ is shown as follows:

\begin{equation}
r^u_{v,i} 
=(\mathbf{x}_v^{U-V})^T \cdot  \mathbf{M} \cdot \mathbf{x}_i^{U-I}.
\end{equation}
Here $(\cdot)^T$ denotes the transposition operator. 
$\mathbf{M}$ is  a learnable metric matrix, harmonizing the embeddings of heterogeneous nodes (video v.s. item).
%
Obviously, the  personalized relevance  score  $r^u_{v,i}$ is dynamically changed with regard to users, videos, and items.

After obtaining the personalized relevance  scores $\{r^u_{v,i}\}$, we normalize them via the softmax function to get the personalized matching scores $s^u_{v\leftarrow i}$ and $s^u_{v\rightarrow i}$, 
shown as follows:
\begin{equation}\label{norm1}
s^u_{v\leftarrow i}=  softmax_i(r^u_{v,i})= \frac{\exp(r^u_{v,i})}{\sum_{i'\in   \mathcal{V}_u^{U-I} }\exp({r^u_{v,i'})}},  
\end{equation}
\begin{equation}\label{norm2}
s^u_{v\rightarrow i}=  softmax_v(r^u_{v,i})= \frac{\exp(r^u_{v,i})}{\sum_{v'\in   \mathcal{V}_u^{U-V}}\exp({r^u_{v',i})}}.
\end{equation}
Note that  the above matching scores are asymmetric since they are normalized on different node sets (i.e., $\mathcal{V}_u^{U-I} $ and $\mathcal{V}_u^{U-V}$).
Given a user $u$,  the personalized matching score $s^u_{v\leftarrow i}$
measures the    matching level from item $i$ to video $v$.  
Intuitively, aggregating relevant items embeddings into video embeddings is able to  the alleviate the  heterogeneity and sparsity problem,  improving user preference understanding and  the effectiveness of video retrieval.



Taking personalized matching scores (e.g., $\{s^u_{v\leftarrow i}\}$)
as coefficients, we can properly summarize all information $ \sum_{i \in \mathcal{V}_u^{U-I}  } r^u_{v\leftarrow i} \cdot   \mathbf{x}_i^{U-I}$ which needs to propagate from relevant items $i \in \mathcal{G}_u^{U-I} $ to video $v$,  and then  utilize it to enhance  the video embedding.
%
%
The final video embedding $\mathbf{h}_v^{U-V}$ of video $v$ is shown as  follows:
\begin{equation}
\mathbf{h}_v^{U-V}  =  \mathbf{x}_v^{U-V} \bigg|\bigg|   \sum_{i \in \mathcal{V}_u^{U-I}  } r^u_{v\leftarrow i} \cdot   \mathbf{x}_i^{U-I}  .
\end{equation}
Also, we   conduct information propagation from    videos to items, and  learn  the final   embedding $\mathbf{h}_i^{U-I}$  of item $i$.
Although node-level graph matching is able to learn the potential connections between videos and items and improves their embeddings via bi-directional information propagation, it only focuses on node-level modeling and largely ignores the user preferences inferred from user interactions.

\begin{table*}[]
\caption{The statistics of the datasets.}
\label{dataset}
\begin{tabular}{|c|c|c|c|c|c|c|c|}
\hline
Dataset                    & Split                  & A-B             & \#A        & \#B        & \#A-B         & \#Sample(Pos)               & \#Sample(All)                \\ \hline
\multirow{4}{*}{7-days}        & \multirow{2}{*}{Train} & User-Video      & 6,980,981  & 899,550    & 129,161,919   & \multirow{2}{*}{14,251,161} & \multirow{2}{*}{71,255,805}  \\ \cline{3-6}
&                        & User-Item       & 6,980,981  & 11,677,996 & 367,232,916   &                             &                              \\ \cline{2-8} 
& \multirow{2}{*}{Val}   & User-Video      & 5,765,456  & 715,610    & 36,464,192    & \multirow{2}{*}{11,346,739} & \multirow{2}{*}{56,733,695}  \\ \cline{3-6}
&                        & User-Item       & 5,765,456  & 5,086,461  & 100,117,118   &                             &                              \\ \hline
\multirow{4}{*}{30-days}       & \multirow{2}{*}{Train} & User-Video      & 18,606,436 & 1,510,227  & 512,317,947   & \multirow{2}{*}{47,346,557} & \multirow{2}{*}{236,732,785} \\ \cline{3-6}
&                        & User-Item       & 18,606,436 & 28,000,470 & 1,478,089,983 &                             &                              \\ \cline{2-8} 
& \multirow{2}{*}{Val}   & User-Video      & 5,765,456  & 715,610    & 36,464,192    & \multirow{2}{*}{11,346,739} & \multirow{2}{*}{56,733,695}  \\ \cline{3-6}
&                        & User-Item       & 5,765,456  & 5,086,461  & 100,117,118   &                             &                              \\ \hline
\multirow{4}{*}{ML-1M} & \multirow{2}{*}{Train} & User-Movie(Cold) & 5,884      & 729        & 123,649       & \multirow{2}{*}{39,912}     & \multirow{2}{*}{154,560}     \\ \cline{3-6}
&                        & User-Movie(Warm) & 6,038      & 3,671      & 630,768      &                             &                              \\ \cline{2-8} 
& \multirow{2}{*}{Val}   & User-Movie(Cold) & 5,884      & 729        & 30,912        & \multirow{2}{*}{7,728}      & \multirow{2}{*}{38,640}      \\ \cline{3-6}
&                        & User-Movie(Warm) & 6,038      & 3,671      & 157,692       &                             &                              \\ \hline
\end{tabular}
\end{table*}

\subsection{Preference-level Graph Matching}


In this section, we introduce a novel method for preference-level graph matching. This method mines the relevance of user preferences for videos and items. Subsequently, this relevance is utilized in cross-graph propagation to refine preference embeddings.


\subsubsection{Extracting User Preferences via Dual Graph Pooling}

Specifically, we propose a novel dual graph pooling method (i.e., $GPN^{U-V}$ and $GPN^{U-I}$) to cluster user-centric sub-graphs (i.e.,  $\mathcal{G}_u^{U-V}$ and $\mathcal{G}_u^{U-I}$) as preference graphs, where each node indicates  a cluster of similar videos or items and  corresponding user preferences.

Given a user $u$ and  $ \mathcal{G}_u^{U-V}$, a graph pooling network   $GPN^{U-V}$ is able to cluster all video embeddings  $\{\mathbf{h}_v^{U-V}   | v \in \mathcal{V}_u^{U-V} \}$ into $k_1$ preferences, and constructs the condensed preference graph $\mathcal{Z}_u^{U-V}$:
%
\begin{equation}
\mathcal{Z}_u^{U-V} = GPN^{U-V}(\mathcal{G}_u^{U-V}),
\end{equation}

\begin{equation}
\{ \mathbf{z}_{u^1}^{U-V},  \cdots, \mathbf{z}_{u^{k_1}}^{U-V} \}= GPN^{U-V}(\mathbf{h}_v^{U-V}  ),
\end{equation}
where $\mathbf{z}_{u^{k_1}}^{U-V} $ is the embedding of  the $k_1$-th user preference.
If we set $k_1 > 1$, then   $GPN^{U-V}$ actually extracts   multiple  types of user preferences. 
Therefore, we simply instantiate $GPN^{U-V}$as the MeanPooling \cite{21www_comb_k}
Similarly, we use    $GPN^{U-I}$ to extract   $k_2$   preference embeddings   $ \{ \mathbf{z}_{u^1}^{U-I},   \cdots, \mathbf{z}_{u^{k_2}}^{U-I} | u^{k_2} \in \mathcal{Z}_u^{U-I} \} $    for user $u$  from $\mathcal{G}^{U-I}$.


\subsubsection{Improving Preference Embedding via Preference-level Graph Matching}

Intuitively, user preferences on both videos and items also have potential relevance. 
So we further propose the preference-level graph matching $GM_{PL}$, which  learns the   preference-level   relevance   matrix  $\mathbf{P}^u \in \mathbb{R}^{k_1 \times  k_2}$ between preference graphs (i.e.,  $\mathcal{Z}_u^{U-V}$ and $\mathcal{Z}_u^{U-I}$), shown as follows:

\begin{equation}
\mathbf{P}^{u}_{} = GM_{PL}(\mathcal{Z}_u^{U-V},\mathcal{Z}_u^{U-I}),
\end{equation}
where each element $p^u_{k_1,k_2} $ indicates the relevance between the $k_1$-th video preference  and  the  $k_2$-th item preference of user $u$,  as follows:

\begin{equation}
p^u_{k_1,k_2} 
=(\mathbf{z}_{u^{k_1}}^{U-V} )^T \cdot    \mathbf{z}_{u^{k_2}}^{U-I}.
\end{equation}

Then, we also normalize  $p^u_{k_1,k_2} $ on $\mathcal{Z}_u^{U-V}$ and $\mathcal{Z}_u^{U-I}$  via softmax, and get corresponding matching scores $s^u_{k_1\leftarrow k_2}$ and $s^u_{k_1\rightarrow k_2}$. Then, the embeddings of preference nodes can be enhanced via bi-directional propagation, as follows:
\begin{equation}
\mathbf{e}_{u^{k_1}}^{U-V}  =  \mathbf{z}_{u^{k_1}}^{U-V} \bigg|\bigg|   \sum_{{u^{k_2}} \in \mathcal{Z}_u^{U-I}  } s^u_{k_1\leftarrow k_2} \cdot   \mathbf{z}_{u^{k_2}}^{U-I}  ,
\end{equation}

\begin{equation}
\mathbf{e}_{u^{k_2}}^{U-I}  =  \mathbf{z}_{u^{k_2}}^{U-I} \bigg|\bigg|   \sum_{{u^{k_1}} \in \mathcal{Z}_u^{U-V}  } s^u_{k_1\rightarrow k_2} \cdot   \mathbf{z}_{u^{k_1}}^{U-V}  .
\end{equation}

Interactively performing graph pooling and graph matching on preference graphs for $L$ times, we can  progressively compress the original graph into  a single  node  (i.e., $k_1=k_2=1$),  termed as $\mathbf{e}_{u}^{U-I}$ and $\mathbf{e}_{u}^{U-I}$,  representing the overall user preferences.

\subsection{Prediction and Optimization}

To comprehensively understand user preferences, we   fuse user preferences on both videos and items, and get 
the final user embedding $\mathbf{z}_{u}$ of user $u$, as follows:

\begin{equation}
\mathbf{z}_{u}= MLP(\mathbf{e}_{u}^{U-V} || \mathbf{e}_{u}^{U-I}|| \mathbf{x}_{u}^{U-V}|| \mathbf{x}_{u}^{U-I}).
\end{equation}
Since the feature of   video  si  much simpler and more stable than   users, we only perform GMN for user modeling,  and the final embedding  $\mathbf{z}_{v}$ of  video  $v$ is actually $\mathbf{x}_v^{U-V}$. Similar to previous works\cite{22cikm_gift,21sigir_surge,23www_high},  we do not use a more sophisticated fusion network (e.g., attention-based fusion), so the effectiveness of bi-level graph matching can be clearly verified.

The estimated preference $\hat{y}_{u,v}$ of  user $u$  towards the target video $v$ is   the   inner product of their embeddings:
\begin{equation}
\hat{y}_{u, v}= (\mathbf{z}_{u})^T \cdot \mathbf{z}_{v}.
\end{equation}

The overall loss function is as follows:
\begin{equation}
\mathcal{L} = \sum_{(u, v, v_{neg}) \in \mathcal{D}}-\ln \sigma\big( \hat{y}_{u,v} - \hat{y}_{u,v_{neg}}\big) + \lambda\cdot ||{\Theta}||^2_2, 
\end{equation}
where $\Theta $ denotes all trainable model parameters, and $\lambda $
controls the L2 regularization strength to prevent overfitting. The whole model can be optimized via back-propagation.

\subsection{Model Analysis}
 Here we give the analysis of the proposed GMN as follows:
 
 \textbullet ~ Considering the   characteristics of video-driven e-commerce, 
 we first represent it as the dual graph, and then reduce user preference 
 understanding to a bi-level graph matching problem. 
 So, the proposed GMN focused on matching relevant nodes (e.g., videos and items) and preferences properly in a bi-level manner, rather than designing better GNN architectures to learn nodes and preferences embeddings.  Similar to previous works\cite{21aaai_hgsrec,21www_comb_k,19kdd_exact_k}, we adopt basic GNNs as backbones, so the effectiveness of  bi-level graph matching  could be clearly verified.

\textbullet ~ We  analyze  the time complexity of GMN in both the node- and preference-level graph matching. In the  node-level graph matching,  if we directly  match all nodes between   $\mathcal{G}^{U-V}$ and $\mathcal{G}^{U-I}$, the computational cost is  $\mathcal{O}(|\mathcal{N}^{V}| \cdot |\mathcal{N}^{I}|)$, which is unacceptable for real-world scenarios. Fortunately, we actually match  user-centric  sub-graphs (i.e.,  $\mathcal{G}_u^{U-V}$ and $\mathcal{G}_u^{U-I}$) , which significantly reduce the time complexity to    $\mathcal{O}(|\mathcal{G}_u^{U-V}| \cdot |\mathcal{G}_u^{U-I}|)$. Similar to the node-level, the    time complexity of preference-level graph matching is     $\mathcal{O}(|\mathcal{Z}_u^{U-V}| \cdot |\mathcal{Z}_u^{U-I}|)$.

\textbullet ~ Then, we further  analyze the space complexity of our GMN, which mainly depends on the learnable parameters in embedding matrixes, rather than deep neural networks. Assuming we have total N nodes and F features in the dual graphs, the space complexity of GMN with feature embedding is  $\mathcal{O}(F \cdot d)$, which is significantly less than ID embedding $\mathcal{O}(N \cdot d)$. In summary, the proposed GMN is able to efﬁciently handle  large-scale e-commerce.

\section{Offline Evaluations}

%
%
%
%


\subsection{Datasets.} 
{Although video-driven e-commerce has grown rapidly on many platforms, related data is protected by privacy   and security policies.
Therefore, we sample   user logs  from a real-world video-driven e-commerce platform, and then extract two large-scale dual graph datasets from different time periods (\emph{7-days}, and \emph{30-days}) for verification.
%
%
%
We   also use a public dataset MovieLens-1M released by GIFT\cite{22cikm_gift} to mimic  video-driven e-commerce, which takes user-movie (warm) as user-item interaction and user-movie (cold) as user-video interaction, respectively.  
%
The details of the datasets  are shown in Table \ref{dataset}.

}

\begin{table*}[]
\centering
\caption{Qantitative results (\%) of different methods. 
Best results  are indicated in bold.  Underlined values (e.g., $\underline{  94.57} $)   indicate that the improvement of GMN is significant based on a  paired \(t\)-test at the significance level of 0.01 .
}

\label{table_auc}
\begin{tabular}{|c|c|cccc|cccc|cccc|}
\hline
\multirow{2}{*}{Type}   & \multirow{2}{*}{Model} & \multicolumn{4}{c|}{7-days}                                                                      & \multicolumn{4}{c|}{30-days}                                                                       & \multicolumn{4}{c|}{MovieLens-1M}                                                                \\ \cline{3-14} 
&                        & \multicolumn{1}{c|}{AUC}    & \multicolumn{1}{c|}{Prec}   & \multicolumn{1}{c|}{Recall} & Loss   & \multicolumn{1}{c|}{AUC}    & \multicolumn{1}{c|}{Prec}     & \multicolumn{1}{c|}{Recall} & Loss   & \multicolumn{1}{c|}{AUC}    & \multicolumn{1}{c|}{Prec}   & \multicolumn{1}{c|}{Recall} & Loss   \\ \hline
\multirow{3}{*}{Single} & DNN             & \multicolumn{1}{c|}{93.03}  & \multicolumn{1}{c|}{81.24}  & \multicolumn{1}{c|}{66.78}  & 24.92  & \multicolumn{1}{c|}{93.16}  & \multicolumn{1}{c|}{80.72}    & \multicolumn{1}{c|}{67.15}  & 24.83  & \multicolumn{1}{c|}{77.53}  & \multicolumn{1}{c|}{73.53}  & \multicolumn{1}{c|}{24.72}  & 41.24  \\ \cline{2-14} 
& MIND                   & \multicolumn{1}{c|}{93.20}   & \multicolumn{1}{c|}{81.44}  & \multicolumn{1}{c|}{67.36}  & 24.62  & \multicolumn{1}{c|}{93.56}  & \multicolumn{1}{c|}{81.22}    & \multicolumn{1}{c|}{68.53}  & 24.11  & \multicolumn{1}{c|}{77.89}  & \multicolumn{1}{c|}{73.63}  & \multicolumn{1}{c|}{24.99}  & 41.04  \\ \cline{2-14} 
& SURGE                  & \multicolumn{1}{c|}{93.30}   & \multicolumn{1}{c|}{81.5}   & \multicolumn{1}{c|}{67.71}  & 24.45  & \multicolumn{1}{c|}{93.65}  & \multicolumn{1}{c|}{81.34}    & \multicolumn{1}{c|}{68.83}  & 23.95  & \multicolumn{1}{c|}{77.93}  & \multicolumn{1}{c|}{73.88}  & \multicolumn{1}{c|}{25.12}  & 40.99  \\ \hline
\multirow{6}{*}{Multi}  & MIND+                  & \multicolumn{1}{c|}{93.51}  & \multicolumn{1}{c|}{81.79}  & \multicolumn{1}{c|}{68.39}  & 24.09  & \multicolumn{1}{c|}{93.99}  & \multicolumn{1}{c|}{81.78}    & \multicolumn{1}{c|}{70.02}  & 23.32  & \multicolumn{1}{c|}{78.18}  & \multicolumn{1}{c|}{74.32}  & \multicolumn{1}{c|}{26.57}  & 40.82  \\ \cline{2-14} 
& SURGE+                 & \multicolumn{1}{c|}{93.61}  & \multicolumn{1}{c|}{81.92}  & \multicolumn{1}{c|}{68.74}  & 23.9   & \multicolumn{1}{c|}{94.02}  & \multicolumn{1}{c|}{81.83}    & \multicolumn{1}{c|}{70.13}  & 23.26  & \multicolumn{1}{c|}{78.19}  & \multicolumn{1}{c|}{74.39}  & \multicolumn{1}{c|}{26.97}  & 40.75  \\ \cline{2-14} 
& GIFT                   & \multicolumn{1}{c|}{93.82}  & \multicolumn{1}{c|}{82.17}  & \multicolumn{1}{c|}{69.46}  & 23.52  & \multicolumn{1}{c|}{94.13}  & \multicolumn{1}{c|}{81.98}    & \multicolumn{1}{c|}{70.49}  & 23.07  & \multicolumn{1}{c|}{78.21}  & \multicolumn{1}{c|}{74.41}  & \multicolumn{1}{c|}{27.67}  & 40.24  \\ \cline{2-14} 
& MGFN                   & \multicolumn{1}{c|}{94.05}  & \multicolumn{1}{c|}{82.47}  & \multicolumn{1}{c|}{70.26}  & 23.08  & \multicolumn{1}{c|}{94.44}  & \multicolumn{1}{c|}{82.43}    & \multicolumn{1}{c|}{71.59}  & 22.46  & \multicolumn{1}{c|}{78.29}  & \multicolumn{1}{c|}{74.45}  & \multicolumn{1}{c|}{28.10}   & 40.01  \\ \cline{2-14} 
& CCDR                   & \multicolumn{1}{c|}{94.12}  & \multicolumn{1}{c|}{82.56}  & \multicolumn{1}{c|}{70.48}  & 22.96  & \multicolumn{1}{c|}{94.66}  & \multicolumn{1}{c|}{82.80}     & \multicolumn{1}{c|}{72.28}  & 22.01  & \multicolumn{1}{c|}{78.35}  & \multicolumn{1}{c|}{74.88}  & \multicolumn{1}{c|}{29.51}  & 39.87  \\ \cline{2-14} 
& GMN                    & \multicolumn{1}{c|}{\textbf{\underline{94.57}}}  & \multicolumn{1}{c|}{\textbf{\underline{83.21}}}  & \multicolumn{1}{c|}{\textbf{\underline{72.03}}}  & \textbf{\underline{22.07}}  & \multicolumn{1}{c|}{\textbf{\underline{95.51}}}  & \multicolumn{1}{c|}{\textbf{\underline{84.27}}}    & \multicolumn{1}{c|}{\textbf{\underline{74.93}}}  & \textbf{\underline{20.19}}  & \multicolumn{1}{c|}{\textbf{\underline{78.65}}}  & \multicolumn{1}{c|}{\textbf{\underline{75.12}}}  & \multicolumn{1}{c|}{\textbf{\underline{30.23}}}  & \textbf{\underline{39.39}}  \\ \hline
/                       & Impro(\%)          & \multicolumn{1}{c|}{1.02} & \multicolumn{1}{c|}{0.78} & \multicolumn{1}{c|}{2.19} & 3.87 & \multicolumn{1}{c|}{1.90} & \multicolumn{1}{c|}{1.78} & \multicolumn{1}{c|}{3.66} & 8.26 & \multicolumn{1}{c|}{1.06} & \multicolumn{1}{c|}{0.32} & \multicolumn{1}{c|}{2.44} & 1.21 \\ \hline
\end{tabular}
\end{table*}

\subsection{Baselines.}
To validate the effectiveness of  GMN, 
we  first select \emph{single} interaction  based methods (i.e., YoutubeDNN\cite{16recsys_youtube_dnn}, MIND\cite{19cikm_mind}, and SURGE\cite{21sigir_surge}) as well as  \emph{multiple} interactions based methods (i.e., GIFT\cite{22cikm_gift}, MGFN\cite{22www_mgfn}, and CCDR\cite{22kdd_ccdr}). 
Since MIND and SURGE cannot be directly applied to the dual graph, we also provide corresponding dual versions (i.e., MIND+ and SURGE+) for better performance. The baselines are shown below:

{


 \begin{itemize}

 \item 
  \textbf{YoutubeDNN}\cite{16recsys_youtube_dnn}: 
It is a classical model for large-scale   video retrieval, utilizing two-tower architecture to learn the embeddings of users and videos based on their features.

 \item \textbf{MIND/MIND+}\cite{19cikm_mind}: 
 MIND is a classical retrieval   model, which   utilizes a dynamic routing mechanism to extract   users' diverse interests from historical behaviors.
 Since MIND cannot directly apply to the dual graph, we extend MIND as MIND+ to learn diverse user interests from both user-video and user-item interactions,  and  then concatenate them as the final user interests for   video retrieval.

 \item 
 \textbf{SURGE/SURGE+}\cite{21sigir_surge}: 
 SURGE is a basic graph-based model that takes graph convolution and graph pooling to extract and fuse users' interest from user  interaction for recommendation. 
 We also extend SURGE as SURGE+ to handle the dual graph. 
 The experimental setting is the same as MIND/MIND+.

 \item 
 \textbf{GIFT}\cite{22cikm_gift}: It is a graph-based   model that utilizes diverse types of relations $r$  to guide the feature propagation between cold- and warm-domain for video  recommendation. 
Here we take   user-video interaction    as cold-domain  and   user-item interaction as warm-domain for experiment.

\item 
 \textbf{MGFN}\cite{22www_mgfn}: It is a multi-graph based video recommendation model which 
 encapsulates interaction patterns across scenarios via  multi-graph fusion network. 

 \item 
 \textbf{CCDR}\cite{22kdd_ccdr}: It is a contrastive cross-domain model for retrieval, which  simultaneously utilizes both intra- and inter-domain contrastive learning. Here we take user-video and user-item as two domains.

 \item \textbf{GMN}: It is our complete model, which performs both node- and preference-level graph matching on the dual graph for video retrieval.

  \end{itemize}

}

%
%
\subsection{Implementation.}
For all models, we random initialize parameters with  Xavier initializer and select Adam as the optimizer.
For a fair comparison, we use
the following hyper-parameters for all models:
%
the dimension of  node embedding is 128, 
the learning rate is 0.0015,
the L2 regularizer is 0.01, and the dropout rate is 0.75. 
For MIND and SURGE, we use the multiple preferences version and set the maximum  number of preferences to 5.
For YoutubeDNN, we use 3 layer DNN (1024-512-256)  for prediction.
For offline evaluation,  we  select  AUC, Precision, Recall, and Loss as evaluation metrics.
We implement all models with TensorFlow 1.15 and run them on  Nvidia  A100 Cluster.
The  code of GMN  and the anoymous version of the  industrial dataset will be released after acceptance.

\subsection{Performance Comparison}~
{ 
The comparison results are shown in Table \ref{table_auc}, where we have the following observations:

\textbullet ~ The proposed GMN consistently performs better than all baselines  with significant gaps.
%
{
Compared to the best performance of baselines, the improvements  of GMN on AUC metric is up to 1.2\%-1.9\% on  large-scale video-driven e-commerce, which indicates the effectiveness of both node- and preference-level  modeling in the proposed GMN for   video  retrieval. 
}

\textbullet ~ Multiple interactions   based methods (i.e., MIND+ and CCDR) outperform single graph methods (i.e., MIND and SURGE). It makes sense because both  user-video and user-item interactions are able to provide valuable information for understanding user preferences.
%
By capturing the correlation  between two types of interaction,
GIFT, MGFN, CCDR, and GMN 
always show their superiority  over 
concatenate based models (i.e., MIND+ and SURGE+).

%

\textbullet ~GNN-based models, including SURGE, GIFT, MGFN, CCDR, and GMN, perform better than traditional DNN-based models (i.e., MIND and YoutubeDNN), indicating the effectiveness of graph structure modeling in the recommendation.

}



\begin{table}[]
\caption{The ablation studies on     graph- and model-level. }
\centering
\label{table_abla}
\begin{tabular}{|c|cc|cc|}
\hline
\multirow{2}{*}{Ablation} & \multicolumn{2}{c|}{7-days}        & \multicolumn{2}{c|}{30-days}       \\ \cline{2-5} 
& \multicolumn{1}{c|}{AUC}   & Prec  & \multicolumn{1}{c|}{AUC}   & Prec  \\ \hline
$GMN_{  \backslash  U-V} $   & 93.09  & 81.29 & 93.2    & 81.55 \\ \hline
$GMN_{  \backslash  U-I}$    & 93.12  & 81.31 & 93.33   & 82.03 \\ \hline
$GMN_{  \backslash  NL}$    & 93.96  & 82.36 & 94.13   & 82.58 \\ \hline
$GMN_{  \backslash  PL}$     & 94.16  & 82.62 & 94.33   & 82.85 \\ \hline
GMN                       & \textbf{94.57}  & \textbf{83.21} & \textbf{95.51}   & \textbf{84.27} \\ \hline
\end{tabular}
\end{table}

\subsection{Ablation Study}
In this section, we conduct both model-level and graph-level ablation studies for further verification, shown in Table~\ref{table_abla}.

\textbullet ~ \textbf{Model-level.}
{ 
We first conduct a model-level ablation study to show how  the  delicate designs in GMN affect   performance. 
As shown in Table \ref{table_abla}, we test two variants of GMN (GMN$_{  \backslash  NL}$ and GMN$_{  \backslash PL}$), which remove node- and preference-level graph matching, respectively. 
%
%
Obviously,  GMN performs better than  both GMN$_{  \backslash  PL}$ and GMN$_{  \backslash NL}$), which clearly verifies  the   superiority of node- and graph-level  matching.
%
Note that the degradation of GMN$_{  \backslash NL}$ is   more significant than GMN$_{  \backslash PL}$,   implying that the node-level graph matching contributes more significantly.
%
The reason is that node-level graph matching is able to capture the potential relevance between videos and items and integrate them seamlessly, serving as the basis of preference extraction and preference-level graph matching.

}

\textbullet ~ \textbf{Graph-level.}
We also conduct graph-level ablation study    to verify the effectiveness of   dual graphs modeling, where 
%
GMN$_{  \backslash   U-V}$ and GMN$_{\backslash U-I}$    remove the user-video graph and user-item graph, respectively.
Note that since graph-level ablation studies
are conducted only on
a single graph (user-video or user-item), so they naturally  build upon model-level ablation studies.
In Figure  \ref{table_abla}, we find that  removing either the user-video graph or the  user-item graph fails to comprehensively capture user preference, leading to suboptimal performance.    
It suggests the necessity of dual graph modeling in video-driven e-commerce.

\begin{figure}
\centering

\includegraphics[width=0.999\columnwidth, height=4.5cm]{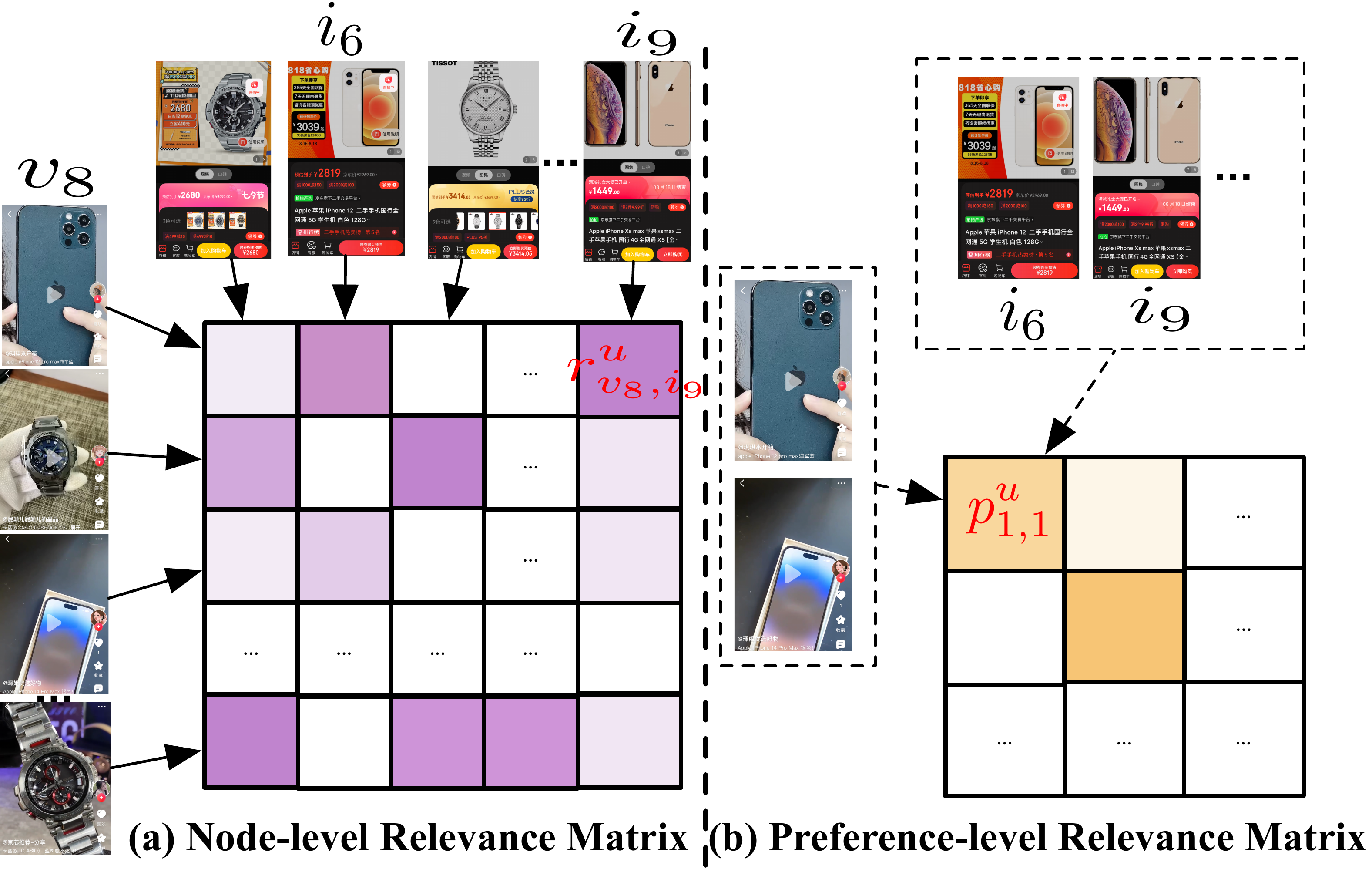}
\caption{
Heatmap on both node- and preference-level relevance  matrix.   Darker color means higher relevance. }
\label{fig_case}
\end{figure}

\subsection{Case Study}
We further present a case study to show the potential interpretability of GMN. 
Taking a young boy as an example who is interested in the Apple iPhone and the Casio watch, he has interacted with related items and videos.
Figure \ref{fig_case} shows the heatmap on both node- and preference-level relevance  matrix, where darker color means higher relevance.
Since item $ i_9$ and video $v_8$  are both related to iPhone,  the node-level relevance score  $r^u_{v_8, j_9}$ is high. A similar phenomenon can be observed at the  preference-level (e.g., $p^u_{1,1}$).

\subsection{Parameter Study}

\begin{figure}
\centering
\subfigure[Dimension of Metric Matrix.]{\includegraphics[width=0.49\columnwidth,height=4cm]{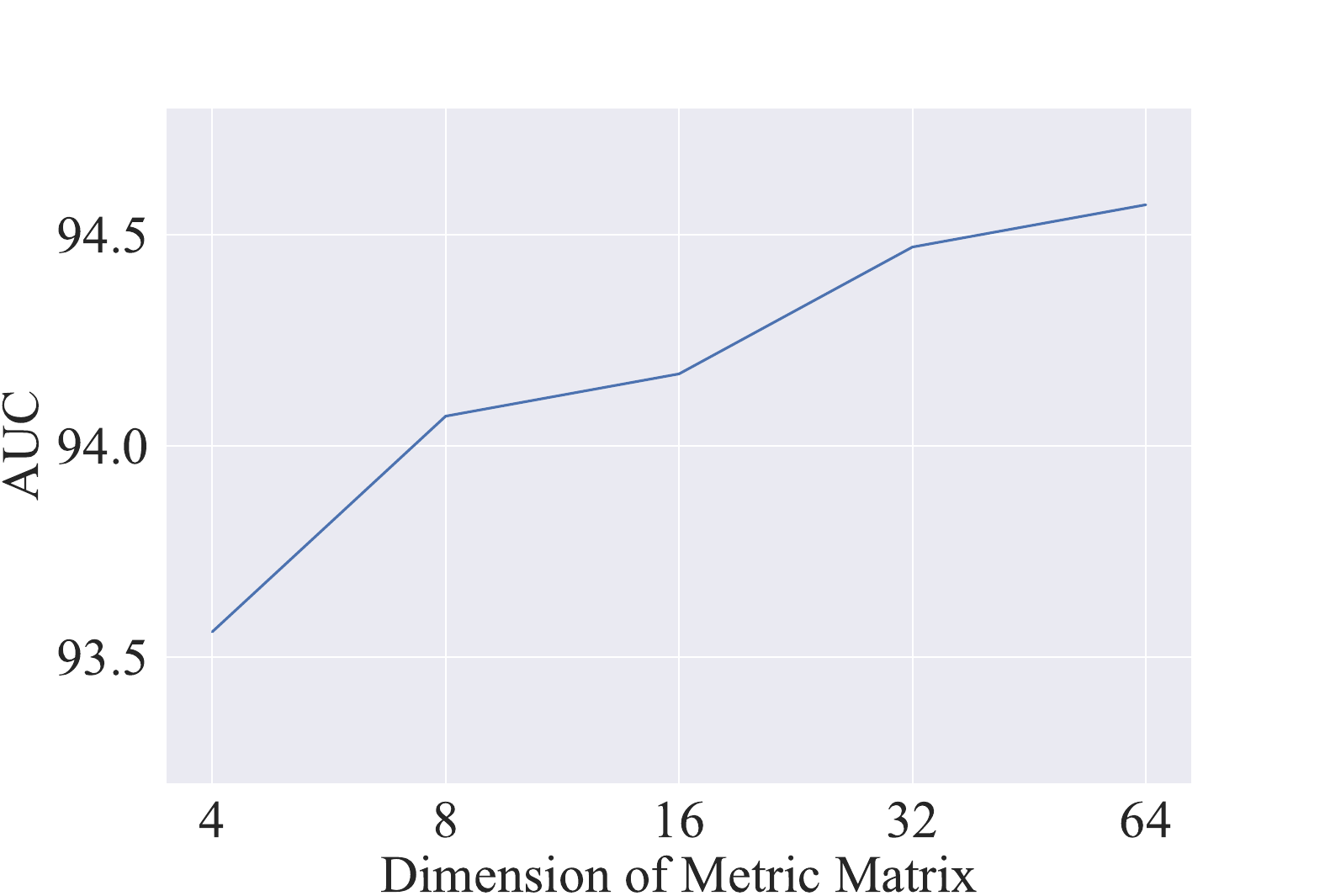}}
\subfigure[Number of User Preferences.]{\includegraphics[width=0.49\columnwidth,height=4cm]{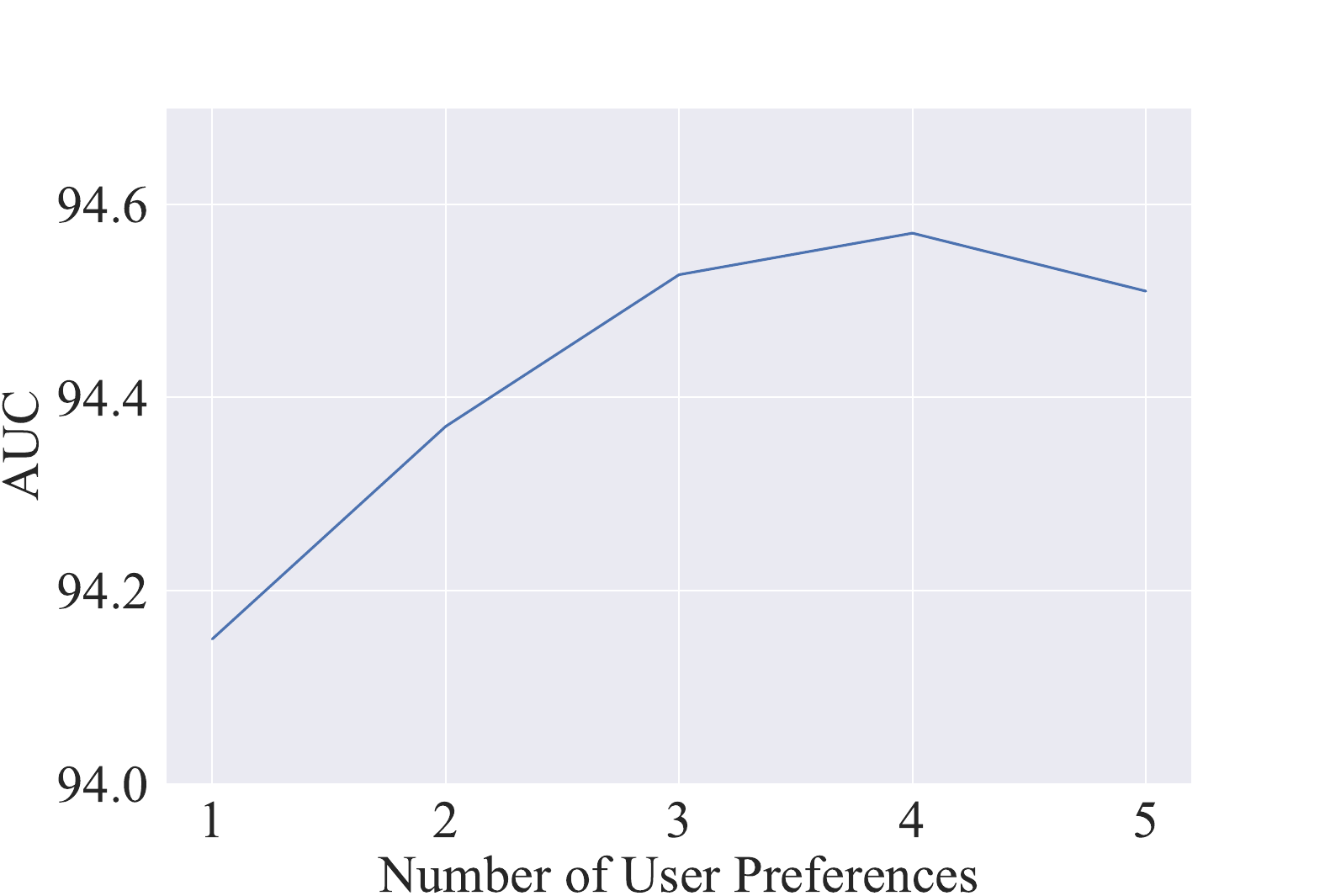}}

\caption{Parameter sensitivity of GMN.}
\label{fig_para}
\end{figure}

 In this section, we investigate the sensitivity of parameters and report the performance of GMN on the offline 7-days dataset with various parameters in Figure \ref{fig_para}.
 
\textbullet~ \textbf{Dimension of Metric Matrix.} As shown in Figure \ref{fig_para}(a),  we  test the effect of the dimension  of the metric matrix $\mathbf{M}$. Obviously, with the growth of the  dimension, the performance of GMN continues to rise. The reason is that, a larger dimension enables the proposed GMN to properly measure the similarity between videos and items.

\textbullet~  \textbf{Number of User Preferences.}   Naturally, most   users have diverse  preferences, and one preference fails to comprehensively describe the characteristics of users. 
Figure \ref{fig_para}(b) shows that  
  more  user preferences  (i.e., larger $k_1$) extracted from  preference-level modeling   improve the overall performance of GMN. 
And, GMN achieves the best performance when  the number of user preferences  is set to 4.

\section{Online Evaluations}

\begin{figure}
\centering

\includegraphics[width=0.9\columnwidth, height=4.5cm]{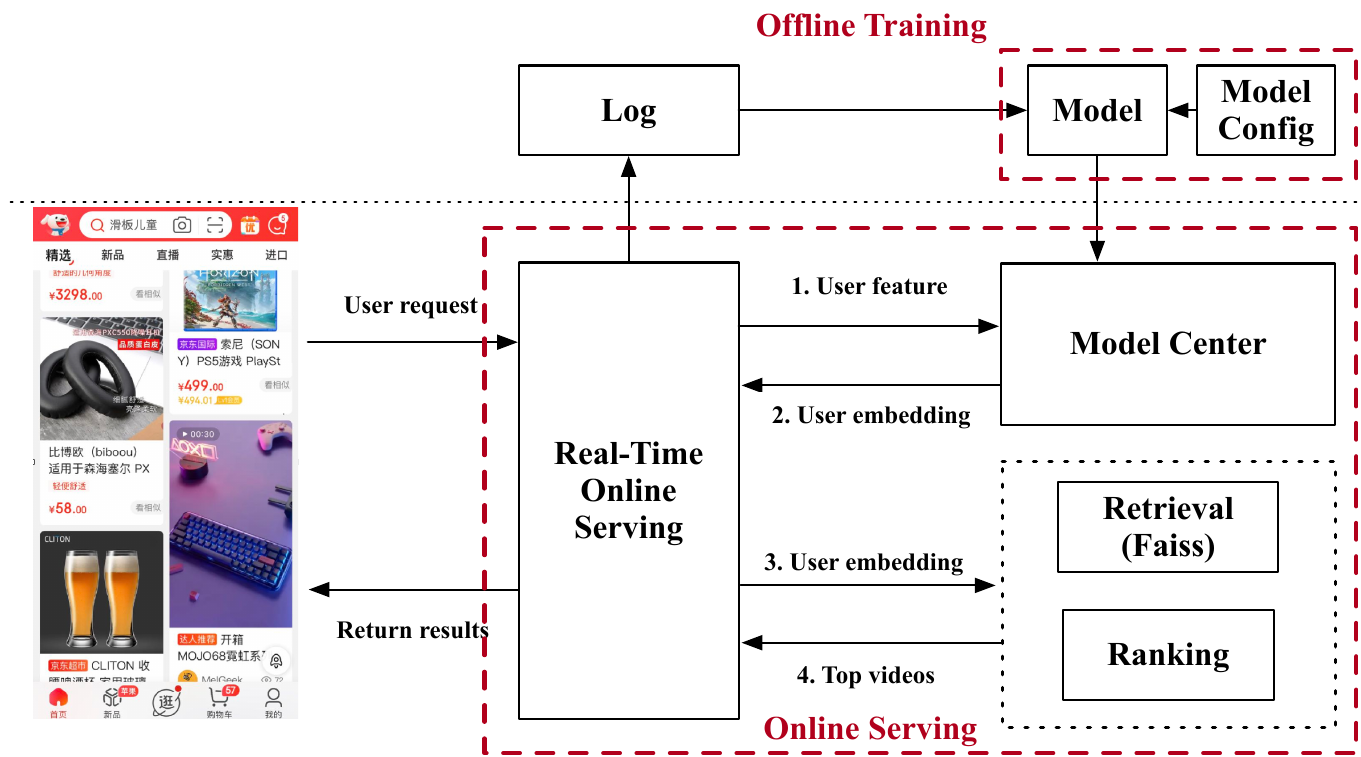}
\caption{
The online system of the proposed GMN. }
\label{fig_arch}
\end{figure}

\subsection{Online A/B Testing}

%

\begin{figure}
\centering
{\includegraphics[width=0.99\columnwidth,height=4cm]{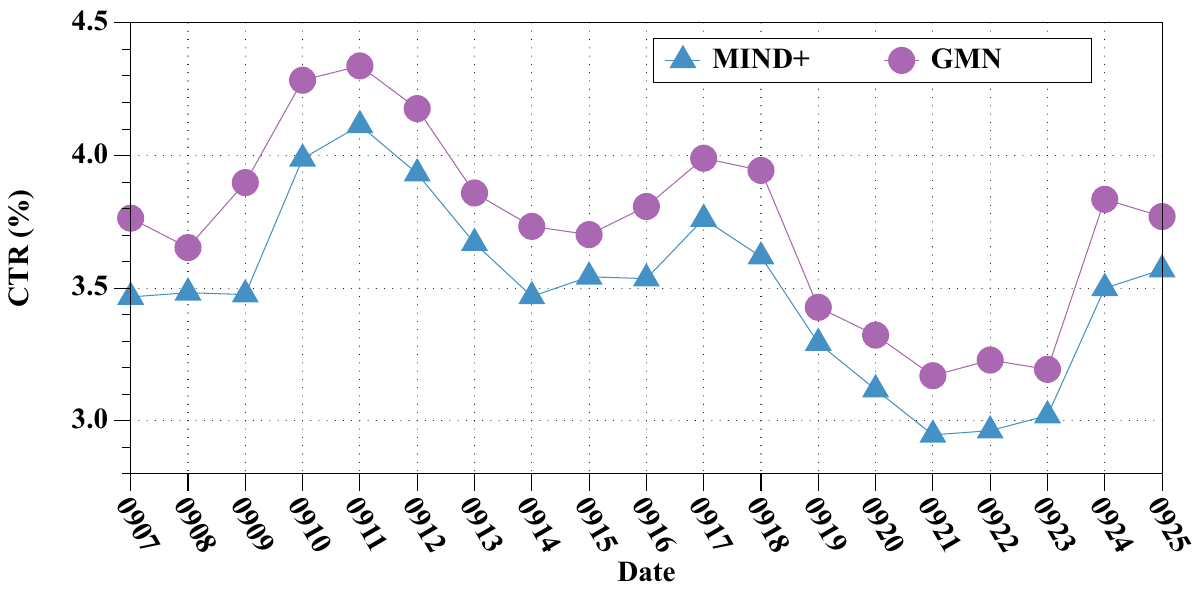}}

\caption{The results of online A/B testing experiments.
}
\label{online}
\end{figure}

\begin{figure}
\centering
{\includegraphics[width=0.99\columnwidth,height=4cm]{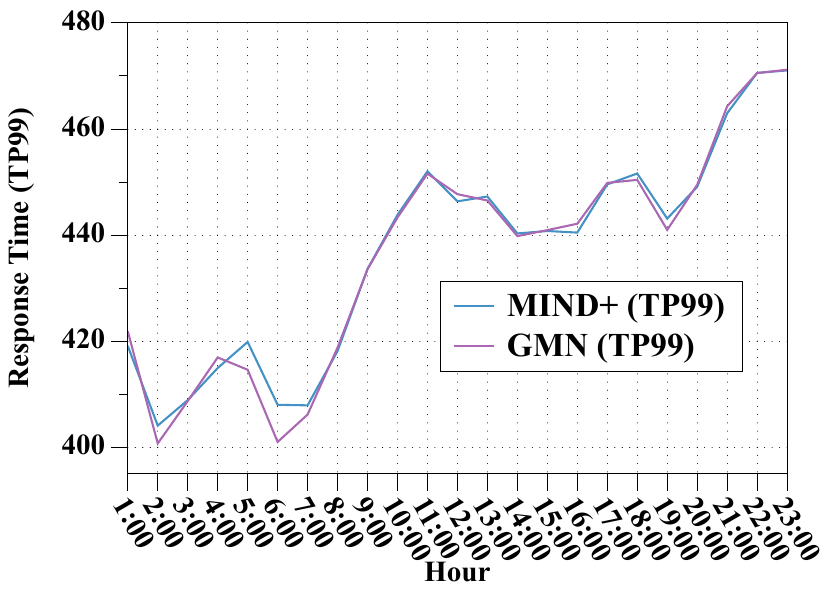}}
\caption{The   response time of online serving.
}
\label{online_time}
\end{figure}

We conduct A/B testing during 0907-0925  (19 days) for online evaluations, shown in Figure \ref{online}. 
Although MIND+ is not the strongest benchmark in offline evaluation, it is the previously deployed model,  affecting hundreds of millions of users every day. Therefore, we use MIND+ as  the   baseline in online A/B testing.
{ 
Here, we select the widely used CTR\footnote{ $CTR=\frac{\text { Number ~of~ clicked ~videos }}{\text { Number~ of~ exposed~ videos }} \times 100(\%)$}  to evaluate   online performance.
%
The larger CTR means users are willing to click the exposed videos recommended  by our system and indicates better performance. 

{
As shown in Figure \ref{online}, the proposed GMN  consistently shows significant superiority over the previously  deployed MIND+ on all days.
} 
More specifically,  the average  CTR of the proposed GMN is  3.74\% while MIND+ achieves only 3.49\%, resulting in a CTR improvement of up to 7.15\%.
%
Following \cite{22cikm_gift}, we also use  TP99\footnote{TP99 is a minimum time that 99\% of queries have been served.} response time to 
evaluate the efficiency of our system, shown in Figure~\ref{online_time}.
%
%
{ 
}
%
%
%
{
Full-day observation shows that the response time of the proposed  GMN system is only about 1 ms higher than the previously developed MIND+,
meeting response time requirements and ensuring user experience.
}
}
Currently, we have deployed GMN on a well-known video-driven e-commerce platform, affecting hundreds of millions of users every day.

%
%
%
%

%
%
%

\subsection{System Architecture and Online Serving.} 
Here we present how to deploy the proposed GMN system for real-world video-driven e-commerce, shown in Figure \ref{fig_arch}.
For offline training,  we collect real-world user clicks as positive samples and random sample several videos as negative samples.   Then, we sample user-centric sub-graphs and associate them with training samples.  as training data. 
The trained model will be divided into two parts and developed in different platforms: (1) The user part, which is developed to the model center, will generate user embedding based on real-time user behavior. (2) The video part, which is developed to the retrieval center, will retrieve candidate videos based on user embedding.

\section{Conclusion}

The thriving of e-commerce is accompanied by a newly emerging paradigm,  video-driven e-commerce,  which utilizes attractive videos to improve user experience and stimulate consumption.
In this paper, we propose the dual graph to represent video-driven e-commerce and innovatively reduce the user preference understanding problem to a graph matching problem. 
To solve it, we further propose a novel  Graph Matching Network (GMN),   which mainly consists of node- and preference-level graph matching.
By properly matching the heterogeneous interactions and preferences, the proposed GMN achieves significant improvements over the state-of-the-art models in both offline and online evaluation (e.g., AUC+1.9\% and  CTR+7.15\%).
Currently, the proposed GMN  has been developed in a well-known video-driven e-commerce platform, affecting hundreds of millions of users every day.

\newpage

\bibliographystyle{ACM-Reference-Format}
\bibliography{ref}

\end{document}